\documentclass[letterpaper, 10 pt, conference]{ieeeconf} 
\IEEEoverridecommandlockouts
\usepackage{url}
\usepackage{cite}
\usepackage{amsmath,amssymb,amsfonts}
\usepackage{graphicx}
\usepackage{textcomp}
\usepackage{xcolor}
\usepackage{caption}
\usepackage{float} 
\usepackage{graphbox}
\usepackage{tcolorbox}
\usepackage[export]{adjustbox}
\usepackage{subcaption}
\usepackage{subfiles}
\usepackage{bibentry}
\usepackage{bm}
\usepackage{multirow} 
\usepackage{makecell}
\usepackage{threeparttable}
\usepackage{pifont}
\usepackage{stfloats}
\usepackage[ruled,vlined,linesnumbered]{algorithm2e}
\usepackage{hyperref}

\title{
\vspace{-0.1in}
\LARGE \bf
Direct Contact-Tolerant Motion Planning With Vision Language Models
}
\author{
He Li$^{1}$, Jian Sun$^{1}$, Chengyang Li$^{3}$, Guoliang Li$^{1}$, Qiyu Ruan$^{1}$, Shuai Wang$^{2,\dagger}$, and Chengzhong Xu$^{1,\dagger}$
\thanks{This work was jointly funded by MOST and The Science and Technology Development Fund (FDCT), Macau SAR (File no. 0074/2025/AMJ), National Natural Science Foundation of China (Grant No. 62371444), Shenzhen Science and Technology Program (Grant No. RCYX20231211090206005), TUBITAK-CAS Bilateral Cooperation Program (Grant No. 321GJHZ2025118MI), and Shenzhen Institutes of Advanced Technology Original Exploration Program.}
\thanks{$^{1}$He Li, Jian Sun, Guoliang Li, Qiyu Ruan, and Chengzhong Xu are with the State Key Laboratory of Internet of Things for Smart City (SKL-IOTSC), University of Macau, Macau, China.}
\thanks{$^{2}$Shuai Wang is with the Shenzhen Institutes of Advanced Technology (SIAT), Chinese Academy of Sciences, Shenzhen, China.}
\thanks{$^{3}$Chengyang Li is with the Department of Electrical and Computer Engineering, The University of Hong Kong, Hong Kong, China.}
\thanks{Corresponding author: Chengzhong Xu ({\tt\small 
czxu@um.edu.mo}) and Shuai Wang ({\tt\small s.wang@siat.ac.cn}).}
\vspace{-0.1in}
}

\begin{document}

\maketitle
\thispagestyle{empty}
\pagestyle{empty}

\begin{abstract}
Navigation in cluttered environments often requires robots to tolerate contact with movable or deformable objects to maintain efficiency. Existing contact-tolerant motion planning (CTMP) methods rely on indirect spatial representations (e.g., prebuilt map, obstacle set), resulting in inaccuracies and a lack of adaptiveness to environmental uncertainties. 
To address this issue, we propose a direct contact-tolerant (DCT) planner, which integrates vision–language models (VLMs) into direct point perception and navigation, including two key components. 
The first one is VLM point cloud partitioner (VPP), which performs contact-tolerance reasoning in image space using VLM, caches inference masks, propagates them across frames using odometry, and projects them onto the current scan to generate a contact-aware point cloud.
The second innovation is VPP guided navigation (VGN), which formulates CTMP as a perception-to-control optimization problem under direct contact-aware point cloud constraints, which is further solved by a specialized deep neural network (DNN).  
We implement DCT in Isaac Sim and a real car-like robot, demonstrating that DCT achieves robust and efficient navigation in cluttered environments with movable obstacles, outperforming representative baselines across diverse metrics.
The code is available at: \url{https://github.com/ChrisLeeUM/DCT}.
\end{abstract}

\section{Introduction}

Navigation in complex environments is a fundamental task for autonomous mobile robots. Traditional navigation algorithms assume strict collision avoidance, treating each obstacle as a rigid body that must be completely avoided \cite{zhang2020optimization,kousik2020safe,han2023rda}. 
However, they can fail in cases when no collision-free paths exist, especially in highly cluttered scenarios.

In practice, many obstacles (e.g., curtains, empty boxes) are movable or deformable and can be safely contacted. Therefore, by allowing controlled contact with such objects, a robot can navigate through spaces otherwise blocked, achieving higher efficiency \cite{frank2009real,stilman2005navigation}. 
This navigation problem, known as contact-tolerant motion planning (CTMP), centers on reasoning whether obstacles can be safely contacted \cite{ellis2022navigation, 10057016} and planning paths among such obstacles \cite{frank2014learning, wang2025contact}.

\begin{figure}[!t]
    \centering
    \begin{subfigure}{0.9\linewidth}
        \centering
        \includegraphics[width=\linewidth]{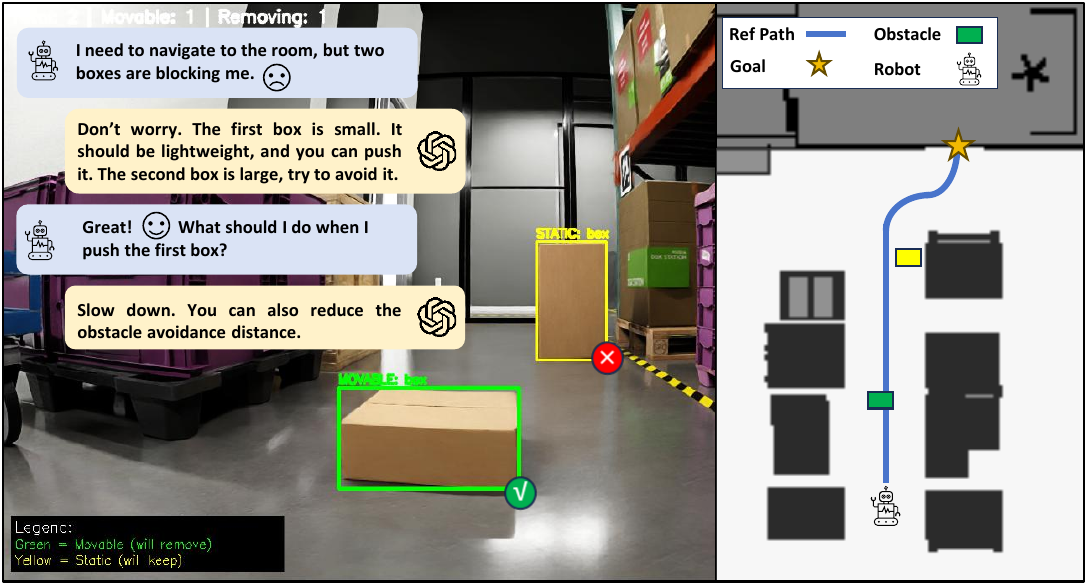}
        \caption{Contact-tolerance reasoning with VLM.}
    \end{subfigure}
    \vspace{0.1in}
    \begin{subfigure}{0.9\linewidth}
        \centering
        \includegraphics[width=\linewidth]{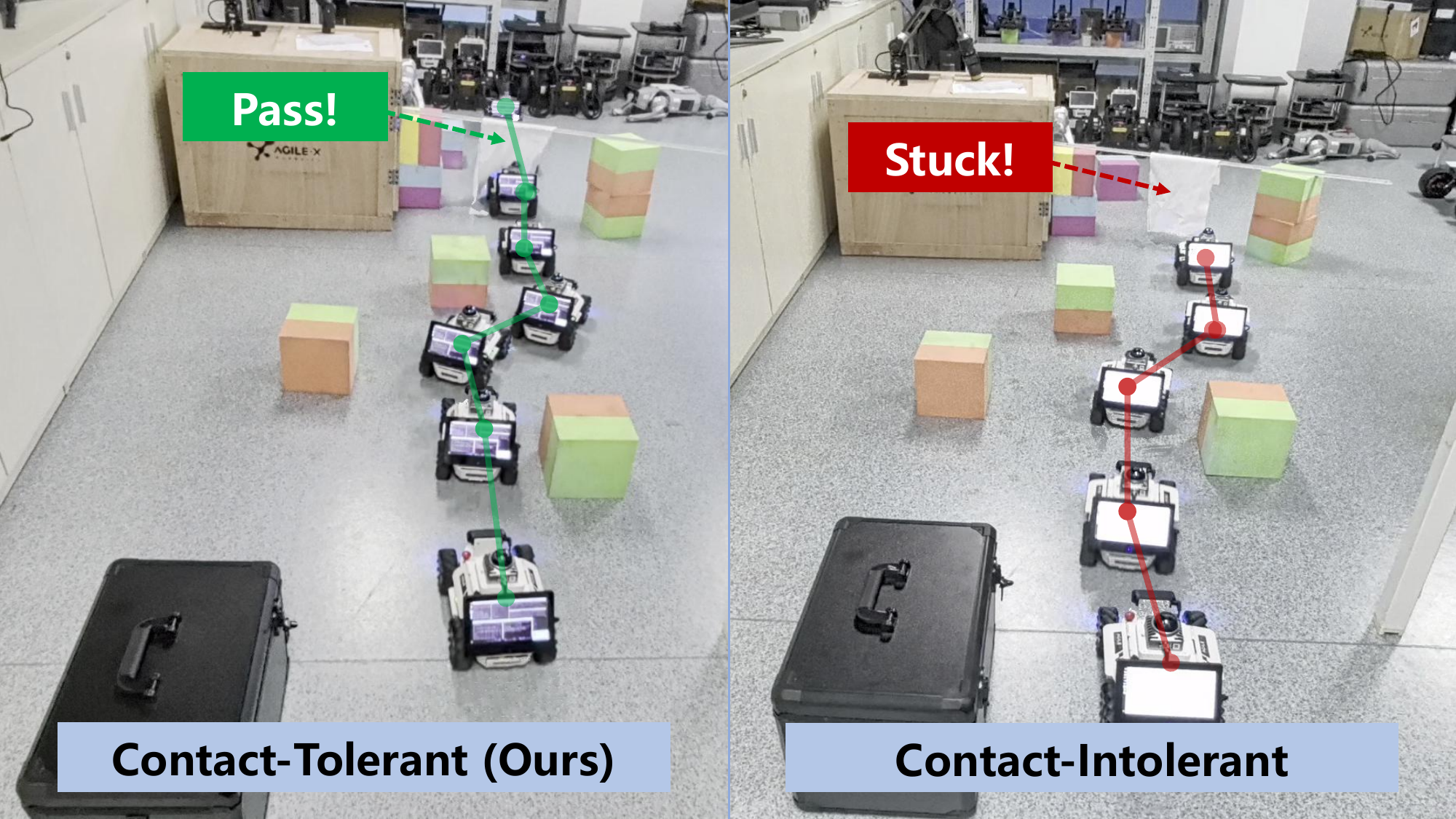}
        \caption{Contact-tolerant vs. contact-intolerant.}
    \end{subfigure} 
    \vspace{-0.15in}
    \caption{DCT motion planning with VLM.}
    \vspace{-0.3in}
    \label{fig:dct_demo}
\end{figure}

Existing CTMP methods can be classified into three paradigms: search-based, reinforcement learning (RL) based, and hybrid planning approaches. 
Search-based methods \cite{Stilman2008,stilman2008planning,zhang2025namo} formulate CTMP as a combinatorial search over obstacle positions and movability cases under all possible interaction sequences prior to execution.
These approaches require prior environmental knowledge and scale poorly as the number of obstacles increases. 
RL-based methods \cite{10057016,yang2025efficient,yao2024local} formulate CTMP as a policy-learning problem, where neural networks optimize action selection through interactions with fixed and movable obstacles. 
These approaches are sensitive to training distribution and often face challenges in real-world deployment. 
Furthermore, how to set appropriate penalty functions for various movable obstacles remains challenging.
Hybrid planning approaches address these limitations by integrating global search with local perception and planning. 
In this paradigm, movable obstacles are typically treated as free or low-cost regions during global search, with their movability assessed dynamically at runtime using visual models \cite{ellis2022navigation}, robot-obstacle contacts \cite{10802169}, or receding-horizon optimization \cite{wang2025contact}. These strategies reduce exhaustive global search and improve flexibility in uncertain environments. 

Despite these advancements in the development of CTMP, existing approaches often rely on indirect spatial representations (e.g., prebuilt map, obstacle set). 
This would inevitably introduce errors (e.g., a convex set cannot match an arbitrary shape) to the planner, leading to conservative policies, or increasing the risk of collisions. 
Using a map may also result in a lack of adaptiveness to environmental changes.
In addition, existing methods struggle with movability reasoning, which is jointly dependent on robot capabilities, obstacle properties, and task-specific requirements.
These issues reduce their robustness and generalizability under real-world uncertainties.
To fill the gap, we propose DCT, a direct contact-tolerant (DCT) motion planning system that leverages vision–language models (VLMs) for contact-tolerant point-to-action navigation, as shown in Fig. \ref{fig:dct_demo}. The DCT consists of two key modules: the \emph{VLM point cloud partitioner (VPP)} and \emph{VPP guided navigation (VGN)}.

The VPP module identifies candidate movable obstacles from RGB images using an open-set detector guided by language prompts, and applies a task-conditioned VLM for mask filtering. 
Since such VLM reasoning cannot be triggered for each point frame due to its inference latency, we propose to cache the (prompt, caption, mask) pair, together with the robot pose obtained using lidar-inertial odometry, into a temporal memory list. 
When the subsequent scan arrives, VPP will retrieve the mask by calling memory from previous VLM executions, propagate the mask across frames through planar homography, and associate the mask with the current lidar scans. This VLM temporal memory design enables the robot to partition each scan into contact-tolerant and contact-intolerant point sets at high frequency. 

The VGN module will adopt the contact-partitioned point cloud to formulate the \emph{direct point distance constraints}. 
This is in contrast to existing CTMP that adopts indirect distance constraints.
However, this would result in a large-scale model predictive control (LMPC) problem with thousands of constraints, which cannot be solved by commercial solvers in real time.
Therefore, we propose a specialized deep neural network (DNN), which trains an agent to solve this LMPC by imitating the optimization process, such that time-consuming iterative computations are converted into real-time feed-forward inference.
In addition, VGN also includes a correcting mode: when pushing fails, contact points will be relabeled as contact-intolerant, and the robot reverses to a safe state for replanning.

Our contributions are summarized as follows:
\begin{itemize}
    \item We propose VPP, a real-time point partitioner that identifies movable objects using VLM contact-tolerance reasoning and memory-based mask propagation.
    \item We propose VGN, a fast learned planner that directly operates on contact-partitioned point cloud and supports real-time robot planning and control.
    \item We implement DCT in both Isaac Sim and real-world environments, with extensive evaluations showing superior navigation performance compared to various baselines in cluttered scenarios.
\end{itemize}

\section{Related Work}
\textbf{Search-based CTMP} treats the problem as a global search task. The planner explicitly reasons over the positions and movability of obstacles to construct a full interaction sequence before execution. Such formulations typically require deciding which obstacles to move, in what order, and how far, yielding an NP-hard combinatorial problem\cite{10.1145/73393.73422}. To solve this problem, existing search-based CTMP methods mostly focus on how to reduce the search or sample complexity. For instance, by constraining each object to a single interaction, it is possible to employ a reverse-order search to reduce the search space\cite{Stilman2008, zhang2025namo, stilman2008planning}.
Moreover, large language models can be adopted to guide non-uniform sampling \cite{zhang2025namo}, enabling robots to efficiently decide which obstacles to move and in what order to push.
However, these methods often rely on a fully observable environment (e.g., map), which may limit their applicability in real-world settings.

\textbf{RL-based CTMP} formulates CTMP as a policy-learning problem, where the neural network learns a policy through interactions with contact-tolerant and contact-intolerant obstacles. Given designed reward functions, the neural network aims to maximize cumulative action rewards of the CTMP task\cite{10057016, yao2024local, li2020hrl4in}. For instance, a curriculum RL framework is utilized to gradually learn from simple collision avoidance tasks to more complex CTMP tasks \cite{10057016}. The designed reward functions enable the policy to make contact with movable obstacles when necessary, while discouraging indiscriminate contact. Moreover, domain randomization enhances policy robustness, enabling effective adaptation to the uncertainties in real-world scenarios \cite{yao2024local}.
Recently, a hierarchical RL framework has been proposed \cite{yang2025efficient}, which enables a mobile manipulator to realize CTMP by combining high-level pushing strategy generation with low-level whole-body control.
Nonetheless, RL-based CTMP approaches suffer from reward sparsity issues and the difficulty of training in real-world environments.

\textbf{Hybrid CTMP} integrates offline global search with online local planning. In this paradigm, movable obstacles are treated as free space or low-cost regions, and their movability is determined during execution. As such, hybrid CTMP alleviates the computational overhead of exhaustive global search\cite{ellis2022navigation, 10802169, wang2025contact}.
For instance,  
extended path planning with pushing interactions has been developed in \cite{ellis2022navigation}, by combining object localization and movability reasoning. 
A directed visibility graph (DV-graph) is introduced \cite{10802169}, where edges encode both traversability and obstacle manipulation costs, and interaction planning updates obstacle affordances for future use.
In \cite{wang2025contact}, an optimization framework is proposed, which models movable objects as interactive elements and incorporates contact dynamics into trajectory optimization, allowing robots to generate smooth, feasible, and predictable paths in cluttered environments.
Nonetheless, existing hybrid methods often rely on indirect spatial representations, which would inevitably introduce errors to the robot actions. 

\section{Problem Statement}

\subsection{Task Description}
We consider a CTMP system, which consists of $1$ autonomous robot, contact-intolerant fixed obstacles $\mathcal{M}^{\mathrm{fix}}$, and contact-tolerant obstacles $\mathcal{M}^{\mathrm{mov}}$.
Given a target $\mathbf{s}^\star$, and robot sensor inputs $\{\mathcal{P},\mathcal{C},\mathcal{I}\}$, where $\mathcal{P}$, $\mathcal{C}$, and $\mathcal{I}$ denote lidar point clouds, camera images, and inertial measurements, respectively, the objective of CTMP is to generate a sequence of robot states $\mathcal{S}$ and executable actions $\mathcal{U}$ reaching the target. 
This can be realized with a repeated receding-horizon planning, where the system only plans a finite sequence of $H$ discretized future steps
$\mathcal{H}_t\triangleq\{t,\cdots,t+H\}$, where $\Delta t$ is the time length between two states.
Specifically, at time $t$, the robot state is denoted as $\mathbf{s}_{t}=(x_{t},y_{t},\theta_{t})$, where $(x_{t},y_{t})$ and $\theta_{t}$ are position and orientation.
The robot action is $\mathbf{u}_{t}=(v_{t},\psi_{t})$, where $v_{t}$ and $\psi_{t}$ are the linear and angular velocities.
The CTMP task at the current horizon needs to generate state sequence
$\mathcal{S}_t=
\{\mathbf{s}_t, \dots, \mathbf{s}_{t+H}\}$ and control sequence
$\mathcal{U}_t=
\{\mathbf{u}_t, \dots, \mathbf{u}_{t+H}\}$ under kinetic constraints 
$\{\mathcal{S}_t,\mathcal{U}_t\}\in\mathcal{F}_t$, with $\mathcal{F}_t$ detailed in Section IV-C.

\subsection{Contact-Tolerant Collision Avoidance}
In CTMP, we need to handle distinct obstacles $\mathcal{M}^{\mathrm{fix}}$ and $\mathcal{M}^{\mathrm{mov}}$, and collision avoidance becomes different from conventional motion planning. 
Particularly, given the current point set $\mathbb{P}_t = \{\mathbf{p}_t^1, \ldots, \mathbf{p}_t^M\}$ obtained from lidar scans, where $\mathbf{p}_t^i \in \mathbb{R}^3$ denotes the position of the $i$-th point and $M$ is the total number of obstacle points, we partition $\mathbb{P}_t$ as $\mathbb{P}_t=\mathbb{P}_t^{\mathrm{mov}} \cup \mathbb{P}_t^{\mathrm{fix}}$, where $\mathbb{P}_t^{\mathrm{mov}}$ denotes contactable points and $\mathbb{P}_t^{\mathrm{fix}}$ denotes contact-intolerant points.
For CTMP, contact with contactable points $\mathbb{P}_t^{\text{mov}}$ is permitted. Therefore, hard collision avoidance should only be applied to $\mathbb{P}_t^{\mathrm{fix}} = \mathbb{P}_t \setminus \mathbb{P}_t^{\mathrm{mov}}$:
\begin{align}
&
{\mathsf{dist}}\left(\mathbb{G}_{t}(\mathbf{s}_t),
\mathbb{P}_t^{\mathrm{fix}} 
\right)
\geq d_{\mathrm{min}},~t\in\mathcal{H}_t, \label{ctmp}
\end{align}
where $\mathbb{G}_{t}$ denotes the robot set and $d_{\mathrm{min}}$ is a predefined safety margin.
To determine the distance $\mathsf{dist}$, it is equivalent to compute the distances between any two points within 
$\mathbb{G}_{t}$ and 
$\mathbb{P}_t^{\mathrm{fix}}$, and then take the minimum:
\begin{equation}
{\mathsf{dist}}\left(\mathbb{G}_{t}(\mathbf{s}_t),
\mathbb{P}_t^{\mathrm{fix}}
\right)
=
\min \{\|\mathbf{x}-\mathbf{y}\| | \mathbf{x}\in\mathbb{G}_{t}, \mathbf{y}\in \mathbb{P}_t^{\mathrm{fix}} \}.  \label{dist}
\end{equation}

\begin{figure*}[!t]
    \centering
    \includegraphics[width=0.98\linewidth]{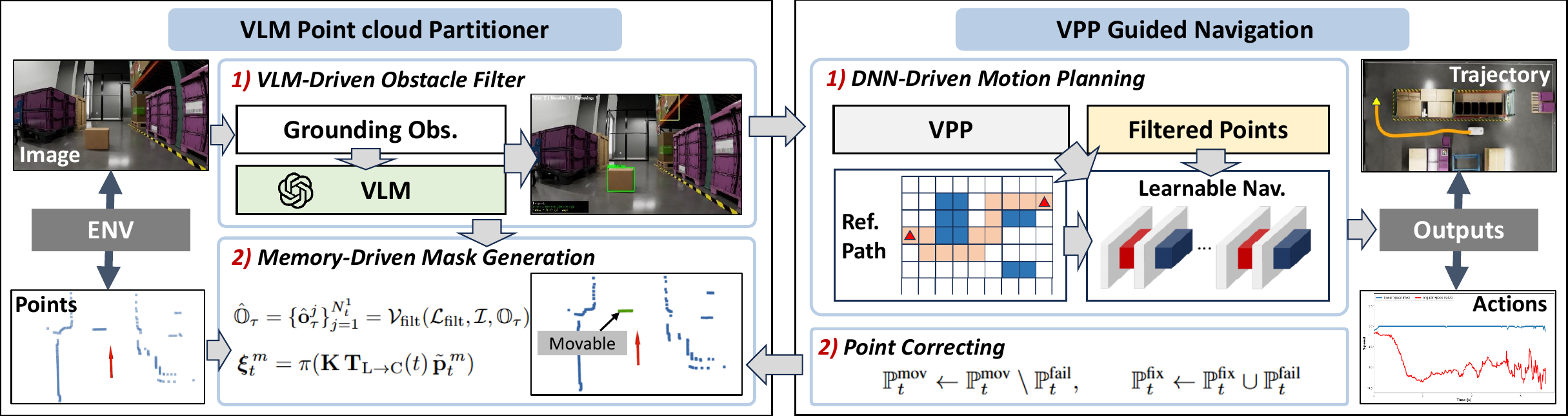}
    \caption{System architecture of DCT, which consists of VPP and VGN modules.}
    \vspace{-0.25in}
    \label{fig:system}
\end{figure*}

\section{Direct Contact-Tolerant Motion Planning}

\subsection{System Overview}

The challenges in tackling CTMP constraint \eqref{ctmp} are two-fold. 
First, the point partition process will rely on other sensor inputs $\{\mathcal{C}, \mathcal{I}\}$, and fusing different data modalities with distinct frequencies is non-trivial.
Second, the number of points in $\mathbb{P}_t^{\mathrm{fix}}$ is large, making direct computation of \eqref{ctmp} computationally intensive. Existing indirect approaches convert $\mathbb{P}_t^{\mathrm{fix}}$ to sets, but this would lead to degradation of solution accuracy.

To address the above challenges, we propose DCT illustrated in Fig.~\ref{fig:system}, which consists of VPP and VGN modules. The VPP module identifies movable objects using VLM contact-tolerance
reasoning and memory-based mask propagation.
It processes raw visual and point cloud inputs, grounds object detections, figures out movable elements, and publishes filtered point representations $\mathbb{P}_{t}^{\mathrm{fix}}$ for downstream planner. 
The VGN module leverages a fast learned planner to map filtered point sets $\mathbb{P}_{t}^{\mathrm{fix}}$ into control signals $\mathbf{u}_t$.
It also consists of a point correction block that updates obstacle movability properties upon pushing failure.

\subsection{Movable Obstacle Grounding with VPP}

The goal of VPP is to identify lidar points corresponding to \emph{movable} obstacles. VLM inference is slow, whereas point cloud filtering requires real-time updates. To address this challenge, we introduce two complementary strategies: VLM-driven filtering, which leverages the reasoning capability of VLMs to accurately categorize obstacles, and real-time mask generation, which enables efficient, high-frequency filtering of lidar data.

\subsubsection{\textbf{VLM-Driven Obstacle Filter}}
At time $t$, given a grounding prompt $\mathcal{L}_g$ and the image $\mathcal{I}_{t}$, we use a grounding model~\cite{liu2024grounding} $\texttt{GRD}$ to localize candidate obstacle regions $\mathbb{O}_{t} = \{\mathbf{o}_{t}^{i}\}_{i=1}^{N_{t}^{0}}$, where ${N_{t}^{0}}$ is the number of detected objects. $\texttt{GRD}$ is an open-set detector that couples vision and language to find objects specified by $\mathcal{L}_g$. Since “movable” depends on task and context (obstacle type, environment, material, and robot capability), we introduce a VLM $\mathcal{V}_{\mathrm{filt}}$ to filter candidates. Conditioned on a task prompt $\mathcal{L}_{\mathrm{filt}}$ together with $\mathcal{I},\{\mathbf{o}_{t}^{N_0}\}$, the VLM returns a movable-obstacle mask list
\begin{align}
    \hat{\mathbb{O}}_{t} = \{\hat{\mathbf{o}}_{t}^{j}\}_{j=1}^{N_{t}^{1}} = \mathcal{V}_{\mathrm{filt}}(\mathcal{L}_{\mathrm{filt}},\mathcal{I}, \mathbb{O}_{t}), 
\end{align}
where $N_{t}^{1}$ is the number of obstacles after filtering. Then we record time $t$ as refresh time $\tau$, the obstacle list $\hat{\mathbb{O}}_{\tau}$, and the corresponding robot position as $\mathbf{s}_{\tau}$.

\subsubsection{\textbf{Memory-Driven Mask Generation}}
\begin{figure}[!t]
    \centering
    \includegraphics[width=0.95\linewidth]{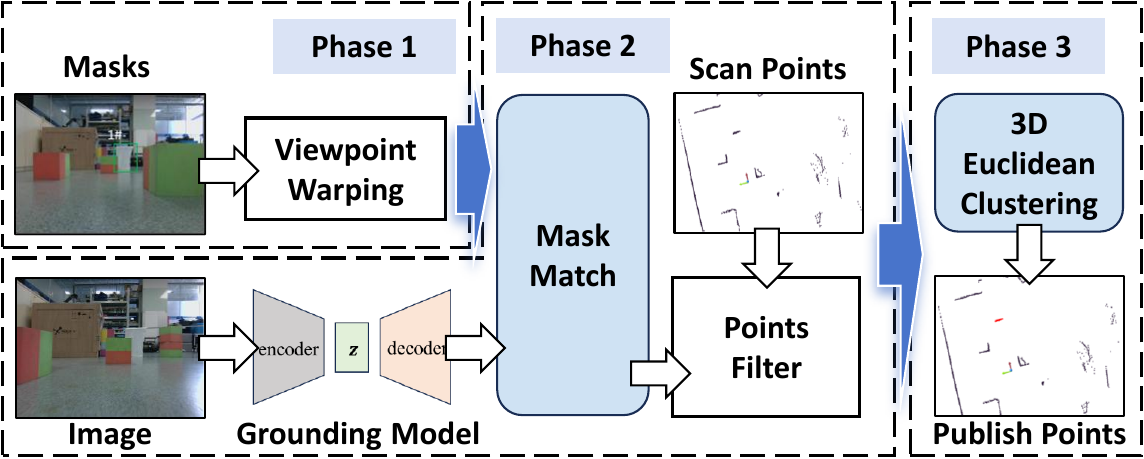}
    \caption{Memory-driven mask generation.}
    \vspace{-0.25in}
    \label{fig:fig_mem}
\end{figure}

Due to limited onboard compute, VLM cannot always be triggered for each point frame. In DCT, VLM detection is triggered only when the moving distance exceeds $d_{\mathrm{thres}}$ or the time since the last detection exceeds $t_{\mathrm{thres}}$. Between two visual inferences, we therefore \emph{reuse and propagate} the last trusted masks and reconcile them when a new detection arrives.  This design will bring challenges to multi-modal data alignment due to the movement of the robot. We address it with a three-phase procedure shown in Fig. \ref{fig:fig_mem}: (i) viewpoint warping; (ii) detection-driven reconciliation; and (iii) point-level refinement and publishing.

\paragraph{Phase I: Viewpoint warping}
Given the robot state at time $t$ and calibrated intrinsics/extrinsics obtained from lidar-inertial odometry, we compute the inter-frame homography $\mathbf{H}_{\tau\rightarrow t}$ and warp each mask:
\begin{equation}
    \hat{\mathbf{o}}_{t}^{\,n} = \mathcal{W}\!\left(\hat{\mathbf{o}}_{\tau}^{\,n},\, \mathbf{H}_{\tau\rightarrow t}\right), \quad
    \hat{\mathbb{O}}_{t}=\{\hat{\mathbf{o}}_{t}^{\,j}\}_{j=1}^{N_{\tau}^{1}},
\end{equation}
where $\mathcal{W}(\cdot)$ applies homogeneous projection to every pixel $(x,y)\!\in\!\hat{\mathbf{o}}_{\tau}^{\,n}$. Because of inevitable state-estimation error, $\hat{\mathbb{O}}_{t}$ may still be misaligned with the current scene.

\paragraph{Phase II: Detection-driven reconciliation}
Given the new image $\mathcal{I}_{t}$, the grounding model yields candidate movable regions
$
\mathbb{O}_{t}=\{\mathbf{o}_{t}^{\,i}\}_{i=1}^{N^{0}_{t}}$.
We match these to the propagated set $\hat{\mathbb{O}}_{t}$ using the intersection-over-union
\begin{equation}
\texttt{IoU}(\mathbb{A},\mathbb{B})=\frac{|\mathbb{A}\cap \mathbb{B}|}{|\mathbb{A}\cup \mathbb{B}|}.
\end{equation}
For each $\hat{\mathbf{o}}_{t}^{\,j}$ in $\hat{\mathbb{O}}_{t}$, define
$i^\star(j)=\arg\max_{i}\ \texttt{IoU}\big(\hat{\mathbf{o}}_{t}^{\,j},\,\mathbf{o}_{t}^{\,i}\big)$,
where $i^\star(j)$ denotes the index of the detection in $\mathbb{O}_{t}$ that attains the highest IoU with the propagated mask $\hat{\mathbf{o}}_{t}^{\,j}$.
If $\texttt{IoU}\!\big(\hat{\mathbf{o}}_{t}^{\,j},\,\mathbf{o}_{t}^{\,i^\star(j)}\big)\ge \sigma_{\mathrm{IoU}}$ (e.g., $\sigma_{\mathrm{IoU}}=0.90$), we regard the pair as a successful match and \emph{update} the trusted mask geometry
$
\hat{\mathbf{o}}_{t}^{\,j}\leftarrow \mathbf{o}_{t}^{\,i^\star(j)}.
$
Propagated masks without a valid match are discarded for safety. Conversely, unmatched detections in $\mathbb{O}_{t}$ are \emph{inserted} as new trusted masks. The result is the reconciled set $\hat{\mathbb{O}}_{t}$.

\paragraph{Phase III: Point-level refinement and publishing}
Let the lidar scan at time $t$ be
\begin{equation}
\mathbb{P}_{t}=\{\mathbf{p}_{t}^{\,m}\}_{m=1}^{M},\ \mathbf{p}_{t}^{\,m}=[x_{t}^{\,m},y_{t}^{\,m},z_{t}^{\,m}]^\top\in\mathbb{R}^3.
\end{equation}
Each point is augmented to $\tilde{\mathbf{p}}_{t}^{\,m}=[x_{t}^{\,m},y_{t}^{\,m},z_{t}^{\,m},1]^\top$ and projected to the image as
\begin{equation}
    \boldsymbol{\xi}_{t}^{\,m}=\pi\!\left(\mathbf{K}\,\mathbf{T}_{\mathrm{L}\rightarrow\mathrm{C}}(t)\,\tilde{\mathbf{p}}_{t}^{\,m}\right), \
    \pi([a,b,c]^\top)=[a/c,\,b/c]^\top,
\end{equation}
with $\boldsymbol{\xi}_{t}^{\,m}\in\mathbb{R}^2$ the pixel coordinates.
Using the reconciled masks $\hat{\mathbb{O}}_{t}$ (let $\hat{\mathcal{U}}_{t}=\bigcup \hat{\mathbf{o}}_{t}^{\,i}$), the scan is partitioned as
\begin{align}
    \mathbb{P}_{t}^{\mathrm{mov}} &= \bigl\{\mathbf{p}_{t}^{\,m}\in\mathbb{P}_{t}\ \big|\ \boldsymbol{\xi}_{t}^{\,m}\in\hat{\mathcal{U}}_{t}\bigr\},\\
    \mathbb{P}_{t}^{\mathrm{fix}} &= \mathbb{P}_{t}\setminus \mathbb{P}_{t}^{\mathrm{mov}}.
\end{align}

While mask gating is effective, image-level decisions ignore 3D continuity and may include isolated speckles or miss nearby returns from the same object. We therefore perform 3D Euclidean clustering (e.g., DBSCAN with radius $\varepsilon_{\mathrm{cluster}}$ and minimum size $N_{\min}$) on $\mathbb{P}_{t}^{\mathrm{mov}}$ to obtain clusters $\mathcal{C}_{t}^{\mathrm{mov}}=\{\mathbb{C}_k\}$. We then apply two symmetric corrections:
\begin{enumerate}
    \item Outlier suppression: points selected by the mask but not belonging to any cluster $\mathbb{C}_k$ with $|\mathbb{C}_k|\ge N_{\min}$ are \emph{removed} from movable:
$
    \mathbb{P}_{t}^{\mathrm{mov}} \gets \bigcup_{k:\,|\mathbb{C}_k|\ge N_{\min}} \mathbb{C}_k.
$
    \item Cluster completion: points in $\mathbb{P}_{t}^{\mathrm{fix}}$ that lie within distance $\varepsilon_{\mathrm{cluster}}$ of some movable cluster $\mathbb{C}_k$ (i.e., $\mathsf{dist}(\mathbf{p},\mathbb{C}_k)\le\varepsilon_{\mathrm{cluster}}$) are \emph{added} to movable:
$
    \mathbb{P}_{t}^{\mathrm{mov}} \gets \mathbb{P}_{t}^{\mathrm{mov}} \cup 
    \bigl\{\mathbf{p}\in\mathbb{P}_{t}^{\mathrm{fix}}\ \big|\ \mathsf{dist}(\mathbf{p},\mathbb{C}_k)\le\varepsilon_{\mathrm{cluster}}, \, \exists\,k \bigr\}.
$
\end{enumerate}
Finally, we publish the refined streams $(\mathbb{P}_{t}^{\mathrm{mov}},\, \mathbb{P}_{t}^{\mathrm{fix}})$ at the lidar rate. This pipeline preserves temporal consistency under sparse visual updates while producing spatially coherent point sets for downstream planning and control.

\subsection{Fast Motion Planning with VGN}

\subsubsection{\textbf{Learned Distance}}
Given $\mathbb{P}_t^{\mathrm{fix}}$ generated by VPP, 
the computation of distance  $\mathsf{dist}$ is a convex optimization problem given by \cite{zhang2020optimization}:
    \begin{equation}
        \begin{aligned}
            \forall \mathbf{p}_t^i\in \mathbb{P}_t^{\mathrm{fix}}: \mathop {{\rm{min}}}\limits_{\mathbf{x}}~ & {\left\| {\mathbf{R}_t(\mathbf{s}_t)\mathbf{x} + \mathbf{t}_t(\mathbf{s}_t) - \mathbf{p}_t^i} \right\|_2^2}
            \\
            \text { s.t. }~                           & ~\mathbf{G}\mathbf{x}{ \preceq }\mathbf{g},
            \label{dis_opt}
        \end{aligned}
    \end{equation}    
where $\mathbf{R}_t({\mathbf{s}_t})$ is the rotation matrix representing the orientation of the robot, $\mathbf{t}_t({\mathbf{s}_t})$ is the translation vector denoting the position of the robot, and matrices $(\mathbf{G},\mathbf{g})$ define the robot edges. 
However, the complexity is high when the number of points in $\mathbb{P}_t^{\mathrm{fix}}$ is large, resulting in unacceptable latency. 

\begin{figure}[!t]
    \centering
    \includegraphics[width=0.95\linewidth]{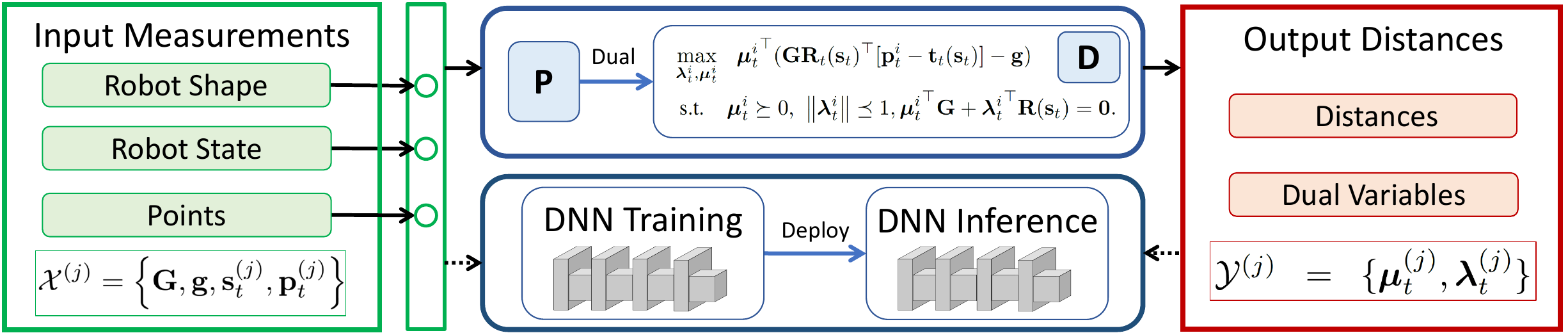}
    \caption{Deep unfolded neural network.}
    \vspace{-0.25in}
    \label{fig:DNN}
\end{figure}

To address this issue, we propose a DNN method for faster distance computation following \cite{han2025neupan}. 
The DNN trains an agent to solve \eqref{dis_opt} by imitating the optimal algorithm, such that time-consuming iterative computations are converted into real-time feed-forward inference. 
The system architecture of DNN is shown in Fig. \ref{fig:DNN}, which consists of three phases.

\paragraph{Phase I: Dual Transformation.}
We first transform problem \eqref{dis_opt} to its dual formulation under the strong duality property\cite{han2023rda,zhang2022generalized}:
\begin{equation}
    \begin{aligned}
         &
         \forall \mathbf{p}_t^i\in \mathbb{P}_t^{\mathrm{fix}}:
         \mathop {\max }\limits_{\bm{\lambda}_t^i, \bm{\mu}_t^i}~~{\bm{\mu}_t^i}^\top ( \mathbf{G}{
\mathbf{R}_t(\mathbf{s}_t)^\top[\mathbf{p}_t^i - \mathbf{t}_t(\mathbf{s}_t)]
         } - \mathbf{g} )                                          \\
         & \text{ s.t. }~~{\bm{\mu}_t^i} \succeq 0, ~ {\left\| \bm{\lambda}_t^i \right\|} \preceq 1, {\bm{\mu}_t^i}^\top\mathbf{G} + {\bm{\lambda}_t^i}^\top\mathbf{R}(\mathbf{s}_t)=\bm{0},
        \label{distance_dual}
    \end{aligned}
\end{equation}
where $\{\bm{\mu}_t^i,\bm{\lambda}_t^i\}$ are dual variables.

\paragraph{Phase II: Demonstration Data Generation.} 
We solve the dual problem \eqref{distance_dual} to generate the demonstration dataset 
$\mathcal{D} = \{\mathcal{A}^{(j)} \}_{j=1}^J$. 
The $j$-th sample is given by $\mathcal{A}^{(j)}   = \{\mathcal{X}^{(j)},\mathcal{Y}^{(j)}\}$, where the input tuple is given by
$$\mathcal{X}^{(j)}=\left\{
\mathbf{G},\mathbf{g},\mathbf{s}_t^{(j)},\mathbf{p}_t^{(j)}
\right\},$$
and the output tuple 
$\mathcal{Y}^{(j)}=\{\bm{\mu}_t^{(j)},\bm{\lambda}_t^{(j)}\}$.

\paragraph{Phase III: DNN Training and Inference.} 
The DNN consists of $6$ fully connected layers with $32$ units, with ReLU, normalization, and tanh activation functions. For model training, we use the mean squared error as the loss function and optimize the network parameters using stochastic gradient descent.
The pre-trained DNN model is deployed at the robot to make real-time inference. 

With DNN, the optimal solution of $\{\bm{\lambda}_t^{i},\bm{\mu}_t^{i}\}$ to \eqref{distance_dual} is obtained in microseconds. 
The distance in \eqref{dist} is thus ${\mathsf{dist}}\left(\mathbb{G}_{h}(\mathbf{s}_h),
\mathbb{P}_h^{\mathrm{fix}} 
\right)=\texttt{DNN}(
\mathbf{G},\mathbf{g},\mathbf{s}_h,\mathbb{P}^{\mathrm{fix}}_{h})$, where $\texttt{DNN}(\cdot)$ takes the minimum of the objective of \eqref{distance_dual} for all $\mathbf{p}_h^i\in \mathbb{P}_h^{\mathrm{fix}}$.

\subsubsection{\textbf{Motion Planning}}
 the VGN problem can be cast as an optimization problem. 
\begin{subequations}
\label{main}
\begin{align}
    \mathsf{P}_t:~&\min_{\mathcal{U}_t,\mathcal{S}_t}\ C_t(\mathcal{S}_t)  \\
    \text{s.t.}~~~&\mathbf{s}_{h+1} = \mathbf{s}_{h} + f(\mathbf{s}_{h}, \mathbf{u}_{h}) \Delta t, ,~h\in\mathcal{H}_t \label{sta_tran}  \\
     ~~~~~~~~~&\mathbf{u}_{\min} \preceq \mathbf{u}_{h} \preceq \mathbf{u}_{\max}, ,~h\in\mathcal{H}_t 
     \label{limit}
     \\
    ~~~~~~~~~&\texttt{DNN}(
\mathbf{G},\mathbf{g},\mathbf{s}_h,\mathbb{P}^{\mathrm{fix}}_{h})
\geq d_{\mathrm{min}},~h\in\mathcal{H}_t, \label{collision}
\end{align}
\end{subequations}
where $C_t(\mathcal{S}_t)$ is the system utility encouraging the robot to approach target $\mathbf{s}^\star$. 
It measures the distance deviation from a given reference path:
\begin{align}
C_t(\mathcal{S}_t) = \sum_{h \in \mathcal{H}_t} \left\| \mathbf{s}_{h+1} - \mathbf{s}_{h+1}^{\diamond} \right\|_2^2,
\end{align}
where
$\mathbf{s}_{h+1}^{\diamond}=\{\mathbf{s}_{t+1}^{\diamond},\cdots,\mathbf{s}_{t+H}^{\diamond}\}$ is the waypoints that are computed using a global path planner (e.g., A*) and $\mathbf{s}^\star$. 
Note that the kinetic set $\mathcal{F}_t$ in Section III now explicitly becomes $\mathcal{F}_t=\{\eqref{sta_tran}, \eqref{limit}, \eqref{collision}\}$, where \eqref{sta_tran} represents the state evolution ($f(\cdot)$ defines vehicle dynamics)
and \eqref{limit} represents the physical constraint \cite{zhang2020optimization} ($\mathbf{u}_{\min},\mathbf{u}_{\max}$ are control limits). 
We collectively define these time-invariant parameters as $\mathcal{P}=\{f, \mathbf{u}_{\min},\mathbf{u}_{\max}, d_{\min}, \mathbf{G}, \mathbf{g}\}$.

This problem can be solved by alternately optimizing $\{\mathcal{U}_t,\mathcal{S}_t\}$ and executing DNN, as shown in \cite{han2025neupan}.
The entire DCT procedure is summarized in Algorithm 1.

\subsubsection{\textbf{Point Correcting}}
VGN enters correcting mode when the robot fails to push through or cross an object. We also treat the robot as stuck if the linear speed is not zero but the pose change between two time steps is very small. Let $\Delta d_t=\|\mathbf{s}_t-\mathbf{s}_{t-1}\|_2$. If $v_t \ge v_{\min}$ and $\Delta d_t \le d_{\text{stuck}}$ for $\tau_{\mathrm{stuck}}$ consecutive steps, the correction is triggered.

First, we update the point labels. We find the points that belong to obstacles that did not move during the failed push and denote them by $\mathbb{P}_{t}^{\text{fail}}$. These points are re-labeled as non-movable. The point sets are then updated as
\begin{equation}
\mathbb{P}_{t}^{\text{mov}} \leftarrow \mathbb{P}_{t}^{\text{mov}} \setminus \mathbb{P}_{t}^{\text{fail}}, 
\quad
\mathbb{P}_{t}^{\text{fix}} \leftarrow \mathbb{P}_{t}^{\text{fix}} \cup \mathbb{P}_{t}^{\text{fail}}.
\end{equation}
Next, the robot moves backward to a previous safe pose. We keep a short history of states $\mathbb{S}^{*}=\{\mathbf{s}_{t},\ldots,\mathbf{s}_{t-F}\}$, where $F$ is the number of traceback steps and $\mathbf{s}_{t-F}$ is the last feasible checkpoint. We temporarily replace the reference path with $\mathbb{S}^{*}$ (traversed in reverse) and execute reverse navigation with speed limits $v_t\in[-v_{\text{rev}},0], v_{\text{rev}}>0$. The robot reverses along the historical states until it reaches $\mathbf{s}_{t-F}$.
Finally, the global planner computes a new reference path $\mathbb{S}^{\text{new}}$ that avoids the newly fixed points, and the robot resumes normal navigation along this path.

\begin{algorithm}[!t]
\caption{\texttt{DCT Motion Planner}}
\KwIn{RGB images $\{\mathcal{I}_{t}\}$, point clouds $\{\mathbb{P}_{t}\}$, robot poses $\{\mathbf{s}_{t}\}$, target $\mathbf{s}^\star$, grounding model $\texttt{GRD}$, task prompt $\mathcal{L}_{\mathrm{filt}}$, model $\mathcal{V}_{\mathrm{filt}}$, thresholds $d_{\mathrm{thres}}, t_{\mathrm{thres}}, \sigma_{\mathrm{IoU}}$, parameters $\mathcal{P}$}

\KwOut{Planned trajectory $\mathbf{s}_{t}$, control sequence $\mathbf{u}_{t}$}

\BlankLine

Memory list $\hat{\mathbb{O}}_{\tau}$; last refresh $(\tau,\mathbf{s}_{\tau})$\;

\For{each time step $t$}{
  \tcp{VLM-Driven Obstacle Filter}
  \If{$\mathrm{dist}(\mathbf{s}_{t},\mathbf{s}_{\tau}) > d_{\mathrm{thres}}$ \textbf{or} $(t-\tau) > t_{\mathrm{thres}}$ \textbf{or} $\hat{\mathbb{O}}_{\tau}=\varnothing$}{
    $\mathbb{O}_{t} \gets \texttt{GRD}(\mathcal{I}_{t})$;
    \\
    $\hat{\mathbb{O}}_{t} \gets \mathcal{V}_{\mathrm{filt}}(\mathcal{L}_{\mathrm{filt}},\mathcal{I}_{t}, \mathbb{O}_{t})$;
    \\
    $(\tau,\mathbf{s}_{\tau}) \leftarrow (t,\mathbf{s}_{t})$, $\hat{\mathbb{O}}_{\tau} \gets \hat{\mathbb{O}}_{t}$;
  }

  \BlankLine
  \tcp{Mem-Driven Mask Generation}
  Warp $\hat{\mathbb{O}}_{\tau}$ to $t$ using pose change\;
  Fresh detection $\mathbb{O}_{t} \gets \texttt{GRD}(\mathcal{I}_{t})$\;
  Reconcile masks by \texttt{IoU} ($\ge \sigma_{\mathrm{IoU}}$);
  \\
  Drop unmatched, obtain $\hat{\mathbb{O}}_{t}$, and update:
  $\mathbb{P}^{\mathrm{mov}}_{t} \gets \mathrm{PointsInside}(\hat{\mathbb{O}}_{t})$,\
  $\mathbb{P}^{\mathrm{fix}}_{t} \gets \mathbb{P}_{t} \setminus \mathbb{P}^{\mathrm{mov}}_{t}$\;
  Refine $\mathbb{P}^{\mathrm{mov}}_{t},\mathbb{P}^{\mathrm{fix}}_{t}$ via clustering\;
  Publish $\mathbb{P}^{\mathrm{mov}}_{t}, \mathbb{P}^{\mathrm{fix}}_{t}$\;

  \BlankLine
  \tcp{ VPP guided navigation}
  $ (\mathcal{S}_t, \mathcal{U}_t)\gets \texttt{VGN}(\mathbf{s}_t,\mathbb{P}^{\mathrm{fix}}_{t}; \mathbf{s}^\star, \texttt{DNN}, \mathcal{P})$\;
}
\end{algorithm}

\section{Experiments}
\begin{figure}[!t]
    \centering
    \begin{subfigure}{0.3\linewidth}
        \centering
        \includegraphics[width=\linewidth]{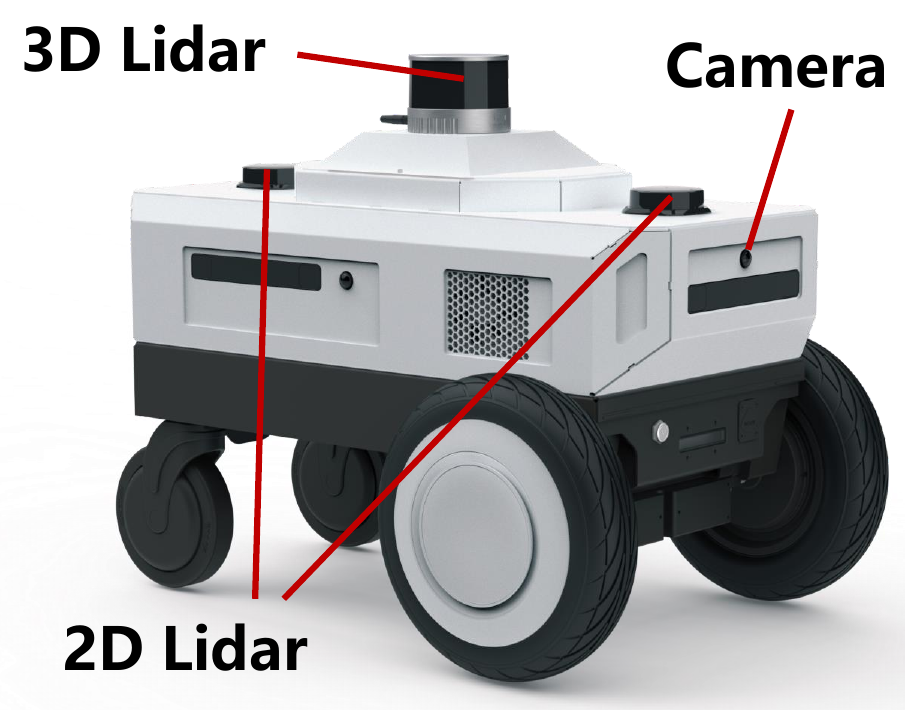}
        \caption{Nova Cater.}
    \end{subfigure}
    \hfill
    \begin{subfigure}{0.64\linewidth}
        \centering
        \includegraphics[width=\linewidth]{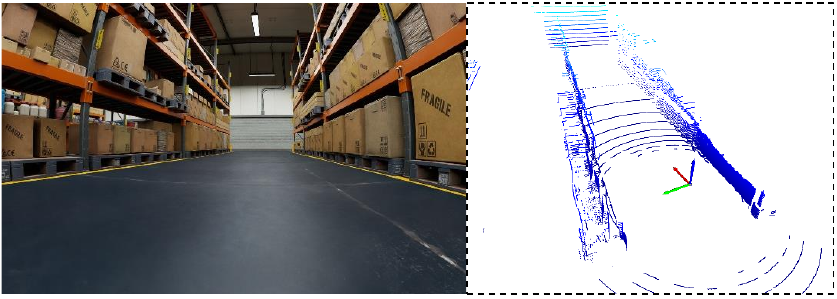}
        \caption{Camera and lidar data.}
    \end{subfigure} 
    \vspace{-0.05in}
    \caption{Simulation setup.}
    \vspace{-0.1in}
    \label{fig:issac}
\end{figure}

We implemented the proposed DCT in Python using ROS Noetic and ROS2. We used Isaac Sim~\cite{isaac_sim_2025}, a high-fidelity simulation platform that provides accurate physical modeling and realistic visual simulation for robot testing and training. The simulation experiment is based on Nova Carter equipped with a simulated lidar and an RGB camera, as shown in Fig.~\ref{fig:issac}. The simulation and ROS packages are implemented on a Linux workstation with an Intel i9-11900 CPU and an NVIDIA RTX 4090 GPU.

\begin{figure}[!t]
    \centering
    \begin{subfigure}{0.45\linewidth}
        \centering
        \includegraphics[width=\linewidth]{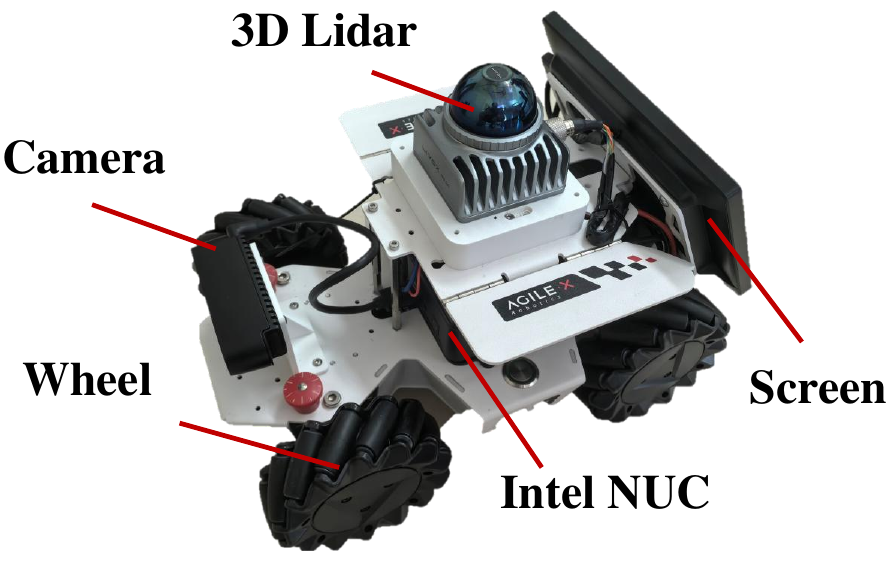}
        \caption{Robot.}
    \end{subfigure}
    \hfill
    \begin{subfigure}{0.26\linewidth}
        \centering
        \includegraphics[width=\linewidth]{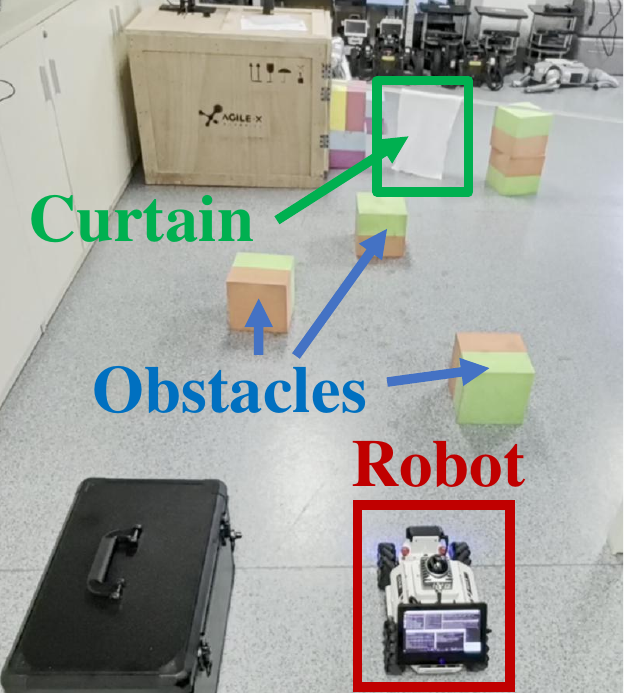}
        \caption{Case 1.}
    \end{subfigure} 
        \hfill
    \begin{subfigure}{0.26\linewidth}
        \centering
        \includegraphics[width=\linewidth]{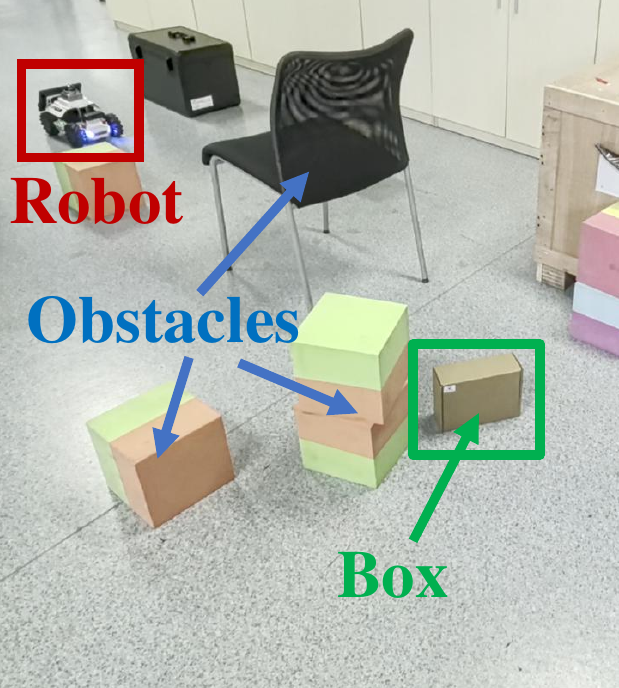}
        \caption{Case 2.}
    \end{subfigure} 
    \vspace{-0.2in}
    \caption{Experimental setup.}
    \vspace{-0.1in}
    \label{fig:exp_setup}
\end{figure}

We also deployed DCT on a real robot in Fig.~\ref{fig:exp_setup}a, with its length, width, longitudinal wheelbase, and lateral wheelbase being 32.2 cm, 22.0 cm, 20.0 cm, and 17.5 cm, respectively.
The robot is equipped with a Livox LiDAR, an RGB-D camera, and an onboard Intel NUC computer, and operates in differential-drive mode.

\subsection{Validation of VPP Under Different VLMs}
In this experiment, we assess the effectiveness of VPP. We construct an evaluation dataset, which consists of image samples annotated with grounding information. The VPP is required to identify all movable or passable obstacles. We selected four state-of-the-art language models for evaluation: GPT-5\cite{openai_gpt5_2026}, Gemini 2.5\cite{google_gemini_2026}, Qwen-vl\cite{qwen_vl}, and Llama 4\cite{touvron2023llama}, and evaluated their performance in terms of Precision, Recall, F1-score, and Accuracy.

\begin{figure}[!t]
    \centering
    \includegraphics[width=0.98\linewidth]{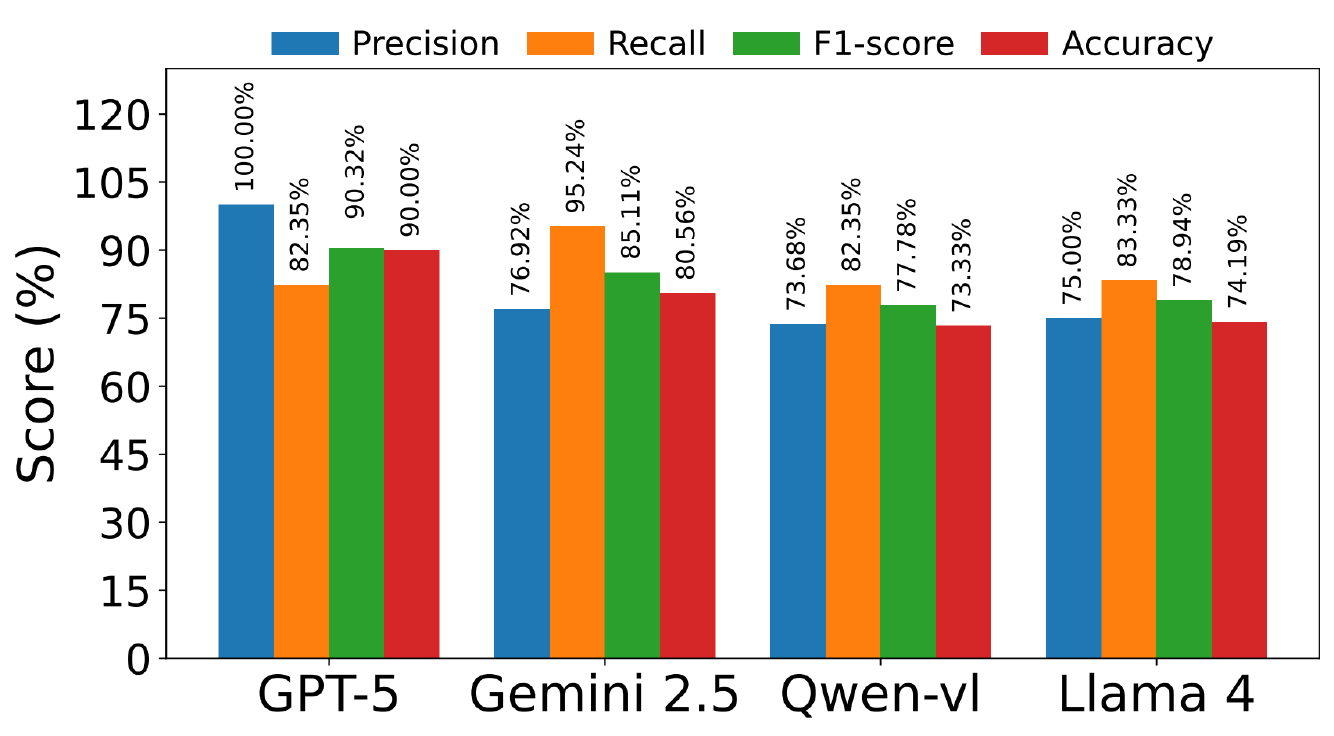}
    \caption{Evaluation of VPP with various VLMs.}
    \label{fig:bar_results}
    \vspace{-0.25in}
\end{figure}

The experimental results of different VLMs are summarized in Fig.~\ref{fig:bar_results}. First, GPT-5 achieved perfect Precision (100\%), demonstrating that all identified pushable obstacles were correct. The remaining three models achieved a Precision of about 75\%, which introduces a potential risk of collision when robots attempt to push incorrectly identified obstacles. This indicates that GPT-5 offers a more reliable choice for robots operating under strict safety requirements.
Second, Gemini 2.5 exhibited the highest Recall (95.24\%), more than 10\% higher than the other models. 
This suggests that Gemini 2.5 adopts a more aggressive selection strategy, achieving higher recall at the cost of lower precision. This may benefit exploration-oriented scenarios where occasional failures are acceptable.
Third, GPT-5 demonstrated the most balanced overall performance. Its F1-score (90.32\%) and Accuracy (90.00\%) significantly exceeded those of Gemini 2.5 (85.11\%, 80.56\%), Qwen-vl (77.78\%, 73.33\%), and Llama 4 (78.94\%, 74.19\%). These results confirm that GPT-5 achieves a balanced trade-off across all metrics, delivering both safety and coverage with high reliability. Thus, GPT-5 was selected for the subsequent experiments.

\subsection{Evaluation in Different Obstacle Scenarios}

To verify the performance gain brought by DCT, we construct three typical obstacle scenarios in Isaac Sim:
\begin{itemize}
    \item \textbf{Case 1:} Movable obstacle with wide side path;
    \item \textbf{Case 2:} Movable obstacle with narrow side path;
    \item \textbf{Case 3:} Fixed obstacle with wide side path.
\end{itemize}
The robot is expected to navigate to the given destination with the obstacle along the route. The target speed is set as 1 m/s for all schemes.
We compare the proposed DCT with two representative approaches: NeuPAN\cite{han2025neupan} and Ellis22\cite{ellis2022navigation}. NeuPAN represents a state-of-the-art direct point navigation method that utilizes lidar measurements to generate a local planned path and accurate control actions.
Ellis22 is a CTMP approach, which incorporates object localization, movability reasoning, and path planning into a unified framework. Notably, Ellis22 requires an occupancy grid map and a clear obstacle set. The experiment results are shown in Table \ref{tab:exp2}.

\begin{figure}[!t]
  \centering
  \subcaptionbox{Case 1 DCT\label{fig:A-a}}[.48\linewidth]{%
    \includegraphics[width=\linewidth]{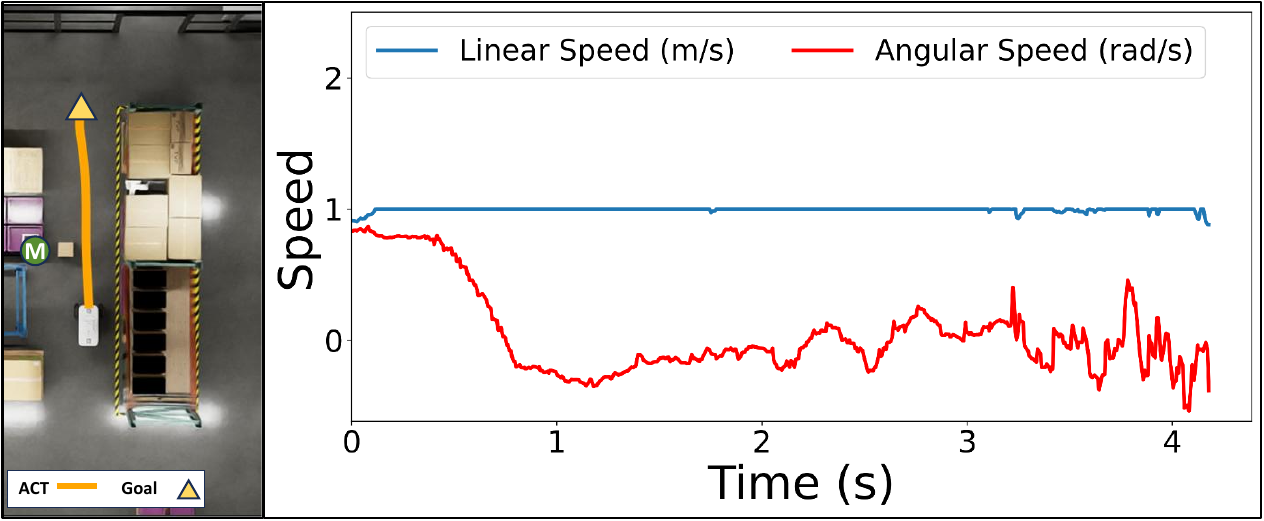}}
  \hfill
  \subcaptionbox{Case 1 Ellis22\label{fig:A-b}}[.48\linewidth]{%
    \includegraphics[width=\linewidth]{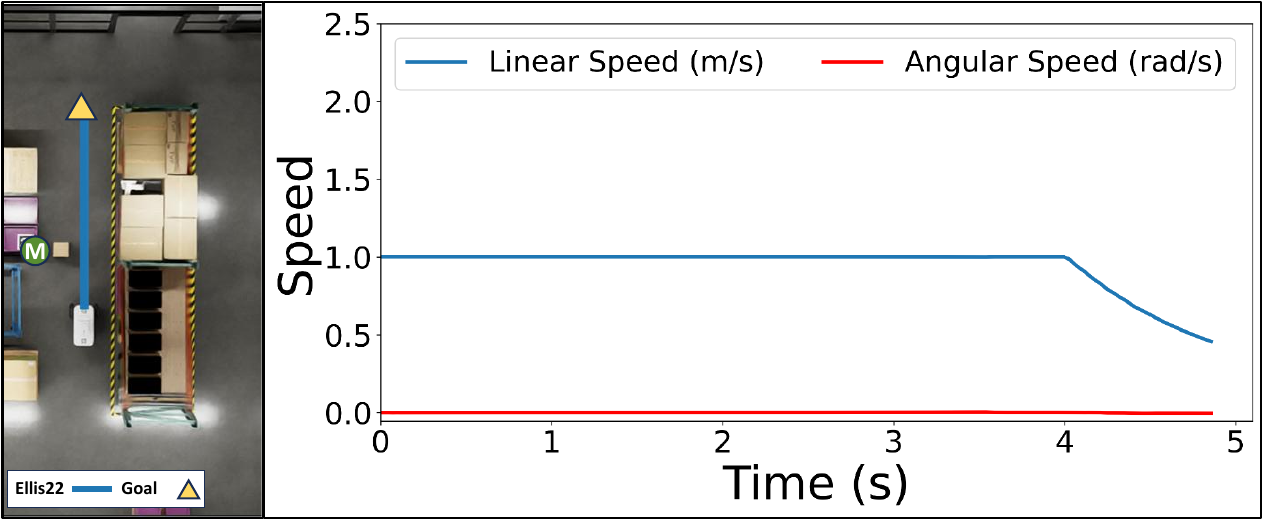}}


  \subcaptionbox{Case 3 DCT\label{fig:A-c}}[.48\linewidth]{%
    \includegraphics[width=\linewidth]{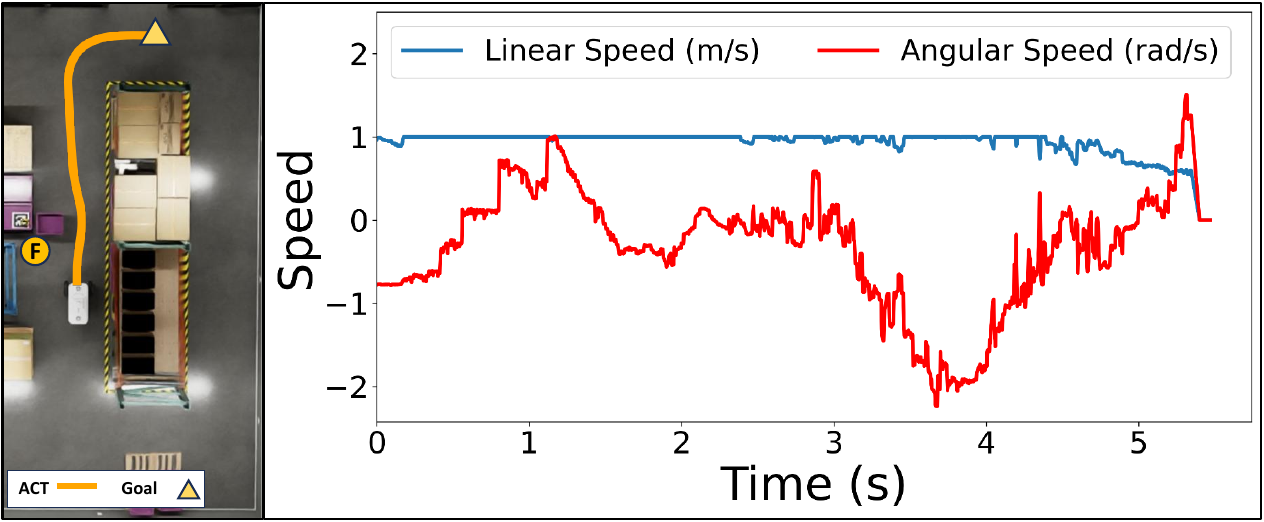}}
  \hfill
  \subcaptionbox{Case 3 Ellis22\label{fig:A-d}}[.48\linewidth]{%
    \includegraphics[width=\linewidth]{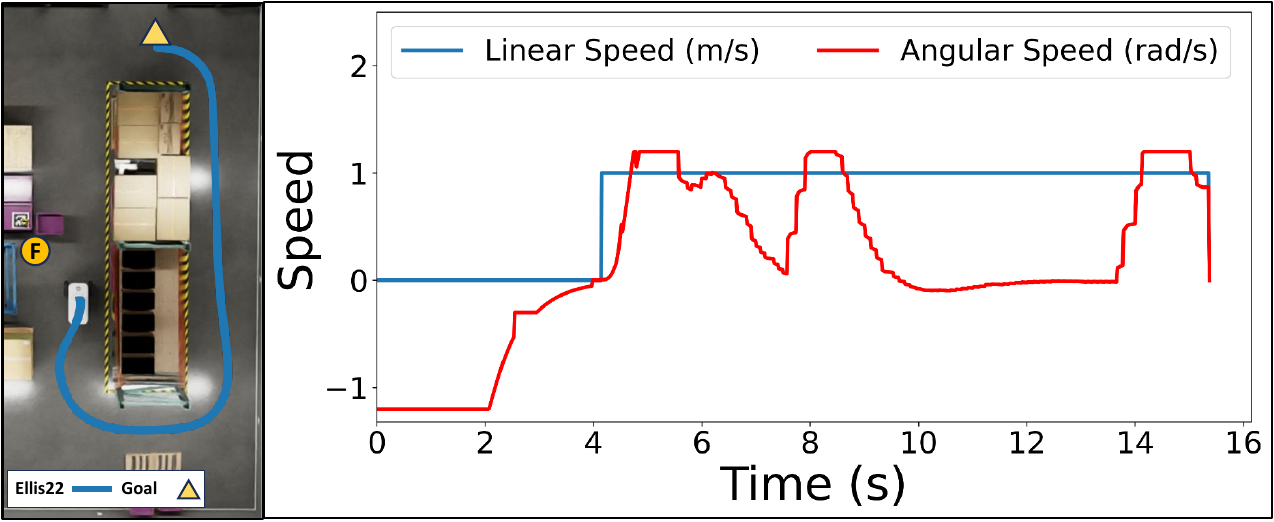}}
  \caption{Evaluation across different obstacle scenarios.}
  \label{fig:exp2}
    \vspace{-0.05in}
\end{figure}

\begin{figure}[!t]
  \centering
  \subcaptionbox{F1M3\label{fig:B-a}}[.48\linewidth]{%
    \includegraphics[width=\linewidth]{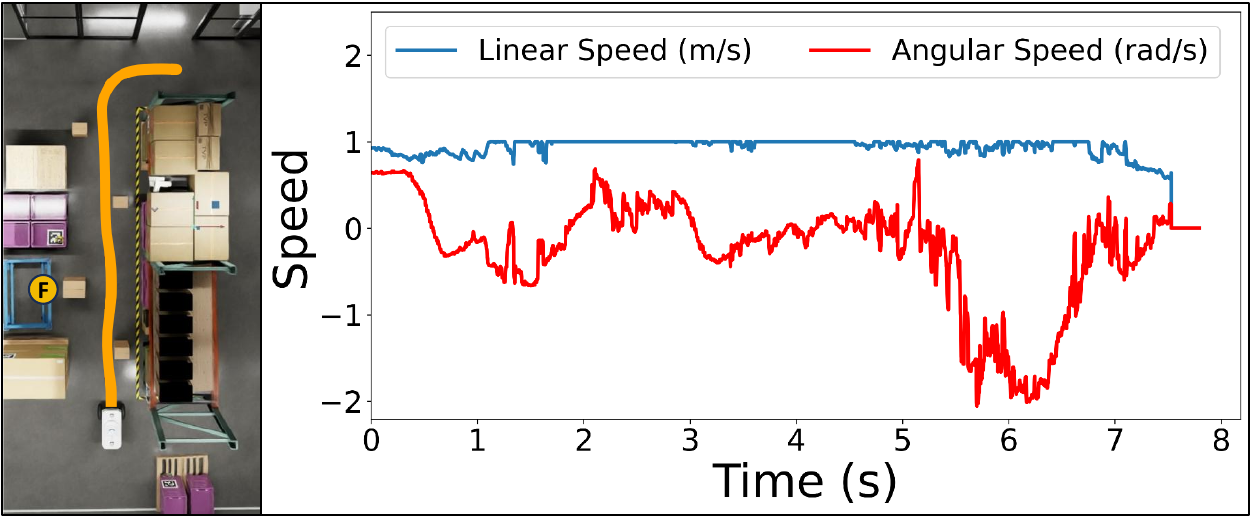}}
  \hfill
  \subcaptionbox{F2M2\label{fig:B-b}}[.48\linewidth]{%
    \includegraphics[width=\linewidth]{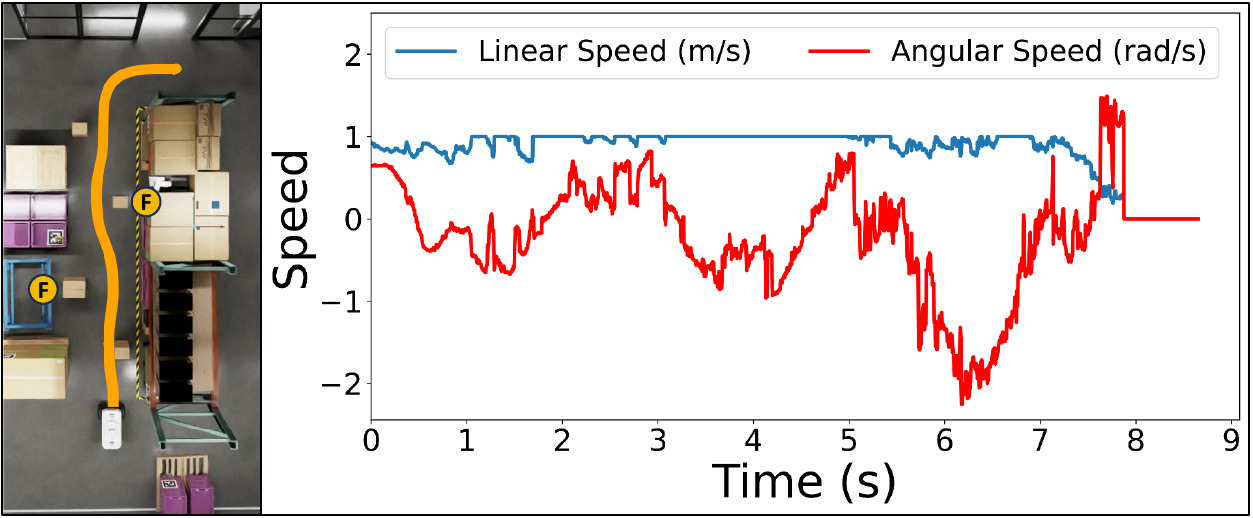}}


  \subcaptionbox{F3M1\label{fig:B-c}}[.48\linewidth]{%
    \includegraphics[width=\linewidth]{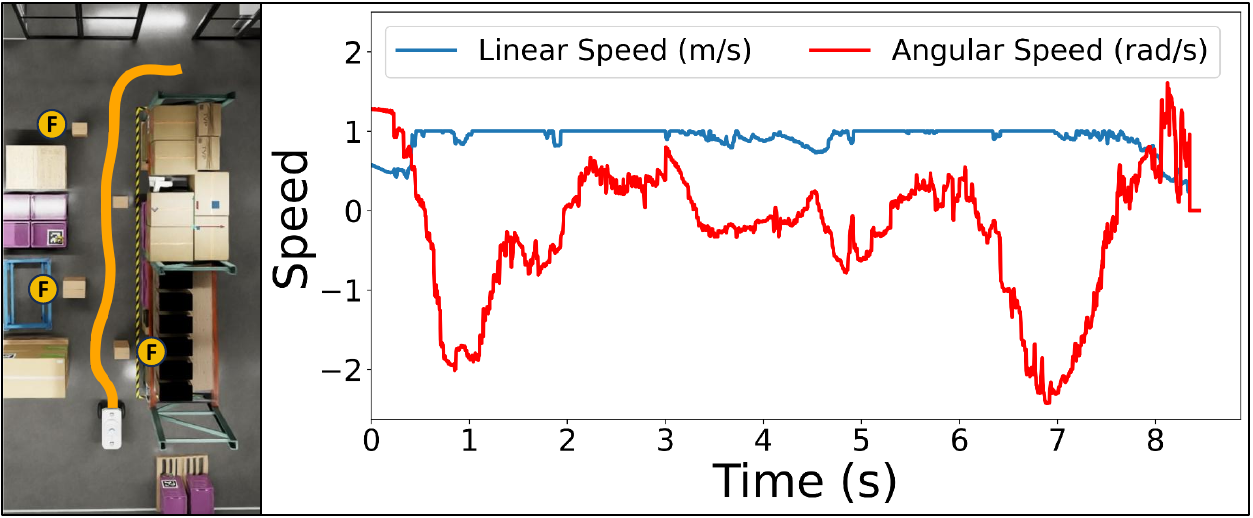}}
  \hfill
  \subcaptionbox{F4M0\label{fig:B-d}}[.48\linewidth]{%
    \includegraphics[width=\linewidth]{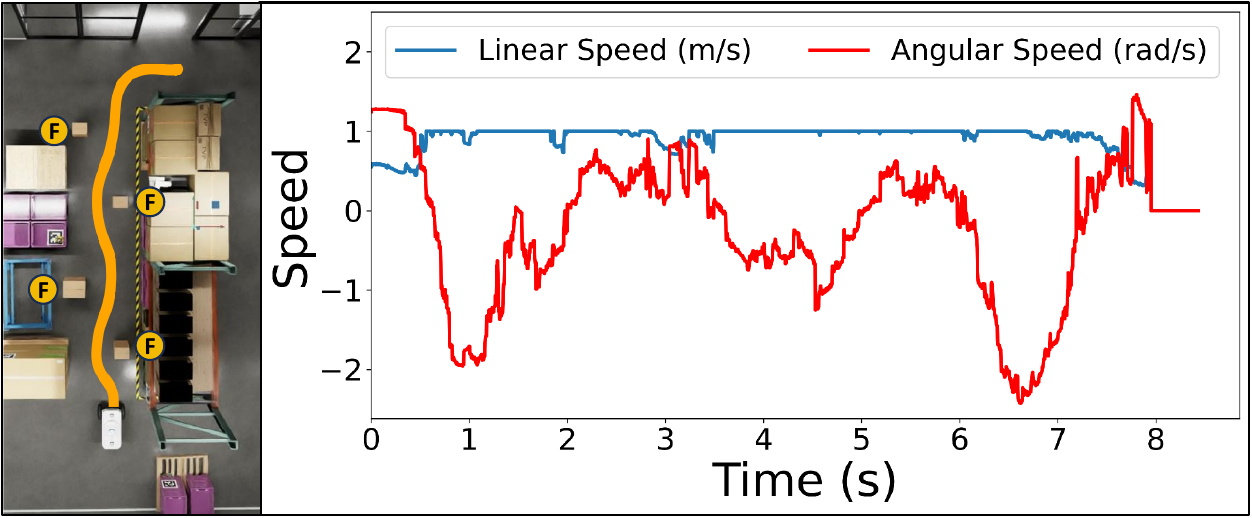}}

  \caption{Evaluation in mixed cluttered environments.}
  \label{fig:exp3}
  \vspace{-0.2in}
\end{figure}

\begin{figure*}[t]
    \centering
    \includegraphics[width=0.98\linewidth]{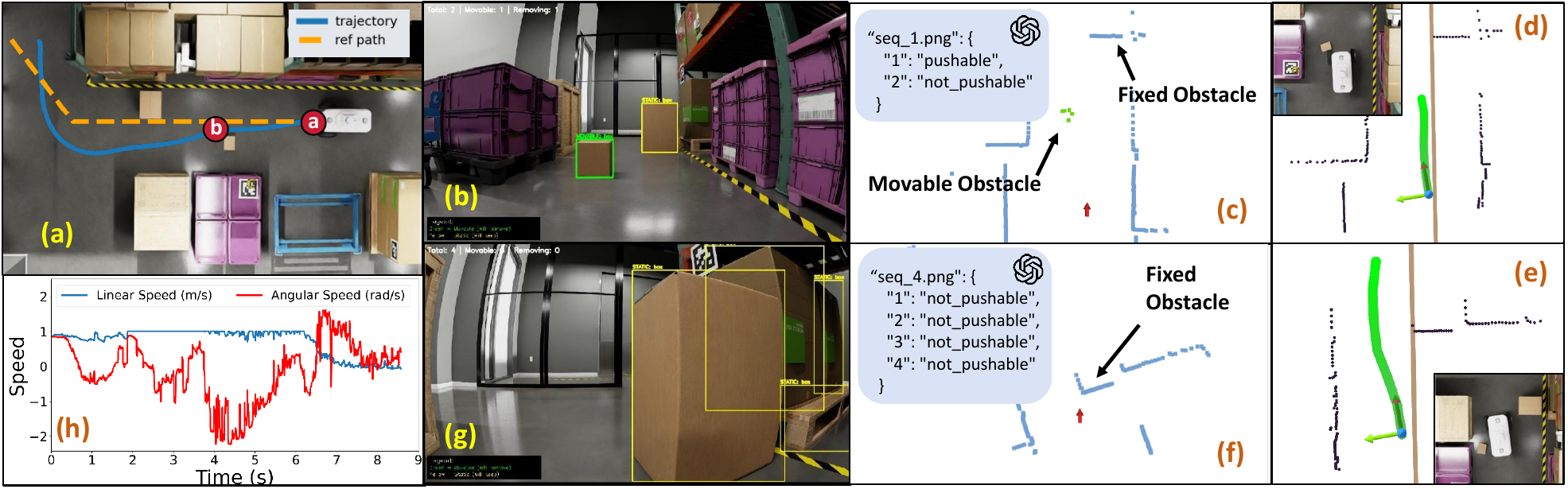}
    \caption{Contact-tolerance reasoning and trajectory-control profiles of DCT in Isaac Sim.}
    \label{fig:exp_demo}
    \vspace{-0.1in}
\end{figure*}

\begin{table*}[t]
    \centering
        \caption{Evaluations in Different Obstacle Scenarios}
        \vspace{-0.05in}
    \resizebox{1.0\linewidth}{!}{
        \begin{tabular}{c ccc ccc ccc ccc}
            \hline
            & \multicolumn{3}{c}{Success}
            & \multicolumn{3}{c}{Navigation Time (s)}
            & \multicolumn{3}{c}{Average Speed (m/s)}
            & \multicolumn{3}{c}{Navigation Distance (m)} \\
            \cline{2-4}\cline{5-7}\cline{8-10}\cline{11-13}
            & DCT & NeuPAN & Ellis22
            & DCT & NeuPAN & Ellis22
            & DCT & NeuPAN & Ellis22
            & DCT & NeuPAN & Ellis22 \\
            \hline
            Case 1 & \ding{51}  & \ding{51}  & \ding{51}  & \textbf{4.22} & 4.37 & 4.917 & \textbf{0.915} & 0.889 & 0.793 & \textbf{3.86} & 3.89 & 3.90\\
            Case 2 & \ding{51}  &   \ding{55}  & \ding{51}  & \textbf{3.54} & - & 4.42 & \textbf{0.938} & - & 0.77 & \textbf{3.22} & - & 3.40\\
            Case 3 & \ding{51}  & \ding{51}  & \ding{51}  & 5.72 & \textbf{5.62} & 15.42 & 0.933 & \textbf{0.950} & 0.749 & \textbf{5.33} & 5.34 & 11.24\\
            \hline
        \end{tabular}
    }
    \vspace{-0.1in}
    \label{tab:exp2}
\end{table*}

In case 1, all three methods successfully reached the goal. DCT achieved the lowest navigation time (4.22s) and the highest average speed (0.915 m/s). In contrast, Ellis22 took the longest navigation time (4.917s) and the lowest average speed (0.793 m/s). This is because Ellis22 adopts the Pure Pursuit for robot control, which prioritizes geometric stability and smooth path tracking, leading to conservative speed control. In contrast, DCT and NeuPAN are trained with efficiency objectives, enabling faster navigation. This can be seen in Fig.~\ref{fig:exp2}a, where the speed of DCT remains almost constant at 1\,m/s, whereas the speed of Ellis22 exhibits a noticeable decrease before reaching the goal position.

In case 2, the side path was intentionally designed to be too narrow to pass without contact. Both DCT and Ellis22 successfully complete the task, while NeuPAN failed to complete the task, as it modeled the movable obstacle as hard constraints and thus could not generate a feasible trajectory. These results underscore the importance of controllable contact in constrained environments and highlight the efficiency advantage of DCT.

In case 3, when facing a contact-intolerant obstacle, DCT, NeuPAN, and Ellis22 showed significantly distinct performance. Ellis22 planned a longer route to avoid the obstacle, resulting in the longest navigation time (15.42s) and distance (11.24m). This can be seen in Fig.~\ref{fig:exp2}d, where the robot first turns around from 0 to 4.2\,s, and then goes around the shelf to reach the end point from 4.2\,s to 16\,s.
This is because the occupancy grid maps of Ellis22 require conservative obstacle inflation, thereby reducing accuracy and limiting applicability in high-precision tasks. In contrast, DCT accomplished the navigation task with significantly shorter navigation time (5.72\,s) and distance (5.33\,m), as shown in Fig.~\ref{fig:exp2}c. These results confirm that DCT adapts well even when obstacles cannot be pushed, maintaining efficiency while being more robust across diverse scenarios.

\subsection{Evaluation in Mixed Cluttered Environments}
To further assess the navigation performance of DCT under mixed cluttered environments, we design scenarios with varying numbers of fixed and movable obstacles. The experiment results are presented in Table \ref{tab:exp3}.

\begin{table}[!t]
    \centering
    \caption{Results in Mixed Cluttered Environments}
    \vspace{-0.05in}
    \begin{tabular}{ccccc}
    \hline
         &  SR&  Nav. Time (s)&  Avg. Speed (m/s)&Nav. Dist. (m)\\
     \hline
         F4M0&  0.7&  8.87&  0.82&7.27\\
         F3M1&  0.9&  8.53&  0.83&7.08\\
         F2M2&  1.0&  7.99&  0.87&6.95\\
         F1M3&  1.0&  7.69&  0.90&6.92\\
     \hline
    \end{tabular}
    \label{tab:exp3}
    \vspace{-0.2in}
\end{table}


The results reveal that in highly cluttered environments composed only of fixed obstacles (F4M0), navigation becomes challenging, with the success rate reduced to 70\% and both navigation time and distance increased.
Changing one obstacle to movable (F3M1) yields a substantial performance gain, achieving a 90\% success rate while lowering navigation cost. When two or more movable obstacles are available (F2M2 and F1M3), the robot consistently achieves 100\% success while also attaining shorter navigation times, higher average speeds, and reduced path lengths. In particular, F1M3 reaches the best performance, with the shortest navigation time (7.69\,s), highest average speed (0.90\,m/s), and shortest path length (6.92\,m). 
The results are corroborated by Fig.~\ref{fig:exp3}. 
When all obstacles are fixed, DCT avoids each one, producing the longest trajectory (Fig.~\ref{fig:exp3}d). Allowing contact with movable obstacles reduces avoidance maneuvers and shortens the navigation distance. In particular, when three obstacles are designated as movable, the navigation distance is markedly reduced, as shown in Fig.~\ref{fig:exp3}a.

Finally, an illustration of DCT in Isaac Sim is shown in Fig.~\ref{fig:exp_demo}.
The result demonstrates that the ability to detect and exploit movable obstacles is crucial in dense environments, as controlled contact improves goal reachability and lowers navigation cost.

\subsection{Verification of DCT on Real Mobile Robot}

\begin{figure*}[t]
\centering
\begin{minipage}[t]{0.48\linewidth}
  \centering
  \includegraphics[width=\linewidth]{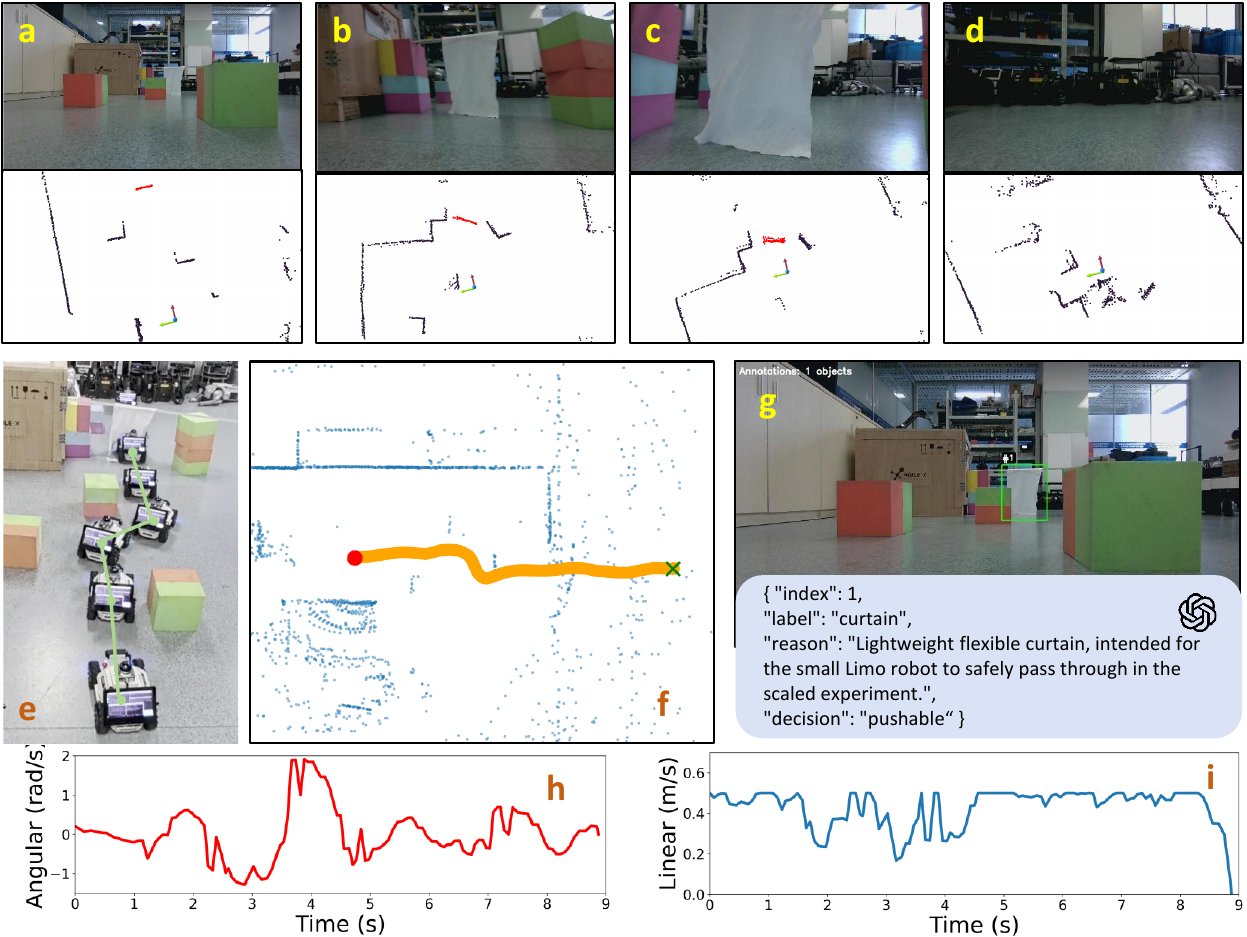}
  \caption{Navigation with deformable obstacle.}
  \label{fig:exp4-curtain}
  \vspace{-0.15in}
\end{minipage}
\hfill
\begin{minipage}[t]{0.483\linewidth}
  \centering
  \includegraphics[width=\linewidth]{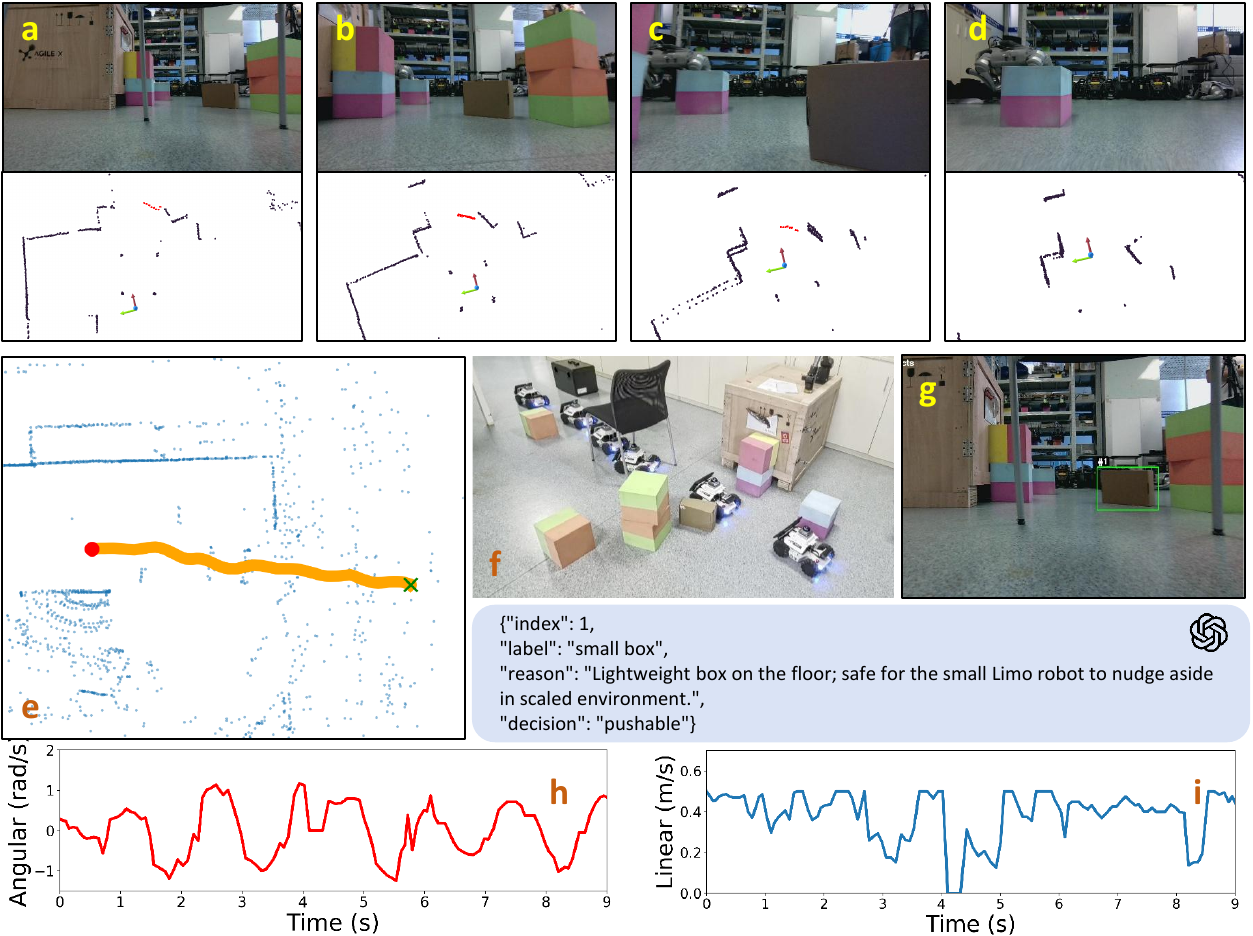}
  \caption{Navigation with movable obstacle.}
  \label{fig:exp4-box}
    \vspace{-0.15in}
\end{minipage}
\end{figure*}

In this experiment, we evaluate the real-world performance of DCT by deploying it on a ground mobile robot in two representative scenarios: (i) traversing a curtain (shown in Fig.~\ref{fig:exp_setup}b) and (ii) navigating among movable boxes (shown in Fig.~\ref{fig:exp_setup}c). 
As shown in Fig.~\ref{fig:exp4-curtain}, in the curtain scenario, DCT identifies the curtain as passable, projects the predicted mask into the scan data, maintains this memory throughout the run, and passes through the curtain to reach the goal position. 
As shown in Fig.~\ref{fig:exp4-box}, in the box scenario, DCT classifies a small box as movable and navigates to the goal position with harmless contact with the small box (Fig.~\ref{fig:exp4-box}), while maintaining hard collision constraints for all other obstacles. 
Interestingly, the robot passes under a chair and avoids the chair legs with agility, which demonstrates its capability to handle arbitrary-shape obstacles under the direct point paradigm.
Across both experiments, DCT restricts physical interaction strictly to obstacles recognized as movable and enforces strict collision avoidance elsewhere, showing its excellent balance between navigation efficiency and safety.

\section{Conclusion}
This paper proposed DCT, a direct contact-tolerant planner for robot navigation among movable obstacles in cluttered environments. 
Through integration of VPP for contact-tolerant reasoning and VGN for contact-tolerant execution, DCT enables the robot to recall short-term memory, make informed contact decisions, act with high accuracy, and maintain efficiency under uncertainty. 
Extensive simulation and real-world evaluations confirm its robustness and superiority over representative baselines. 
Future work will extend DCT toward richer multi-modal reasoning and large-scale real-world deployments.

\bibliographystyle{ieeetr}
\bibliography{main}
\end{document}